\documentclass[sigconf]{acmart}

\makeatletter
\def\@ACM@checkaffil{
    \if@ACM@instpresent\else
    \ClassWarningNoLine{\@classname}{No institution present for an affiliation}%
    \fi
    \if@ACM@citypresent\else
    \ClassWarningNoLine{\@classname}{No city present for an affiliation}%
    \fi
    \if@ACM@countrypresent\else
        \ClassWarningNoLine{\@classname}{No country present for an affiliation}%
    \fi
}
\makeatother

\usepackage{graphicx} 
\usepackage{times}
\usepackage{latexsym}

\usepackage{graphicx} 
\usepackage{algorithm}
\usepackage{algorithmic}
\usepackage{amsmath}
\usepackage{amsfonts}
\usepackage{booktabs}
\usepackage{multirow}
\usepackage{diagbox} 

\usepackage{newfloat}
\usepackage{listings}
\usepackage{hhline}
\usepackage[T1]{fontenc}

\usepackage[utf8]{inputenc}

\usepackage{microtype}

\AtBeginDocument{%
  }

\setcopyright{acmlicensed}
\copyrightyear{2018}
\acmYear{2018}
\acmDOI{XXXXXXX.XXXXXXX}

\acmConference[Conference acronym 'XX]{Make sure to enter the correct
  conference title from your rights confirmation emai}{June 03--05,
  2018}{Woodstock, NY}
\acmISBN{978-1-4503-XXXX-X/18/06}

\begin{document}

\title{Unsupervised Text Representation Learning via Instruction-Tuning for Zero-Shot Dense Retrieval}



\author{Qiuhai Zeng}
\authornote{Those authors contributed equally to this research.}
\authornote{Work done while intern at Amazon.}
\affiliation{%
  \institution{The Pennsylvania State University}
}
\email{qjz5084@psu.edu}
\author{Zimeng Qiu}
\authornotemark[1]
\affiliation{%
  \institution{Amazon AGI}
}
\email{zimengqi@amazon.com}

\author{Dae Yon Hwang}
\authornotemark[1]
\affiliation{%
  \institution{Amazon AGI}
}
\email{dyhwang@amazon.com}

\author{Xin He}
\affiliation{%
  \institution{Amazon AGI}
}
\email{xih@amazon.com}

\author{William M. Campbell}
\affiliation{%
  \institution{Amazon AGI}
}
\email{cmpw@amazon.com}









\begin{abstract}
Dense retrieval systems are commonly used for information retrieval (IR). They rely on learning text representations through an encoder and usually require supervised modeling via labelled data which can be costly to obtain or simply unavailable. In this study, we introduce a novel unsupervised text representation learning technique via instruction-tuning the pre-trained encoder-decoder large language models (LLM) under the dual-encoder retrieval framework. We demonstrate the corpus representation can be augmented by the representations of relevant synthetic queries generated by the instruct-tuned LLM founded on the Rao-Blackwell theorem. Furthermore, we effectively align the query and corpus text representation with self-instructed-tuning. 
Specifically, we first prompt an open-box pre-trained LLM to follow defined instructions (i.e. question generation and keyword summarization) to generate synthetic queries. Next, we fine-tune the pre-trained LLM with defined instructions and the generated queries that passed quality check. Finally, we generate synthetic queries with the instruction-tuned LLM for each corpora and represent each corpora by weighted averaging the synthetic queries and original corpora embeddings. 
We evaluate our proposed method under low-resource settings on three English and one German retrieval datasets measuring NDCG@10, MRR@100, Recall@100. We significantly improve the average zero-shot retrieval performance on all metrics, increasing open-box FLAN-T5 model variations by $[3.34\%, 3.50\%]$ in absolute and exceeding three competitive dense retrievers (i.e. mDPR, T-Systems, mBART-Large), with model of size at least $38\%$ smaller, by $1.96\%$, $4.62\%$, $9.52\%$ absolute on NDCG@10.
\end{abstract}




\maketitle

\section{Introduction}

Dense retrieval systems commonly employ dual-encoder retrieval models which use two separate encoders, either symmetric or asymmetric, to represent the query and corpus \cite{DBLP:journals/corr/abs-1811-08008, DBLP:conf/emnlp/KarpukhinOMLWEC20, DBLP:conf/acl/YangCAGLCAYTSSK20, DBLP:conf/emnlp/DongNBAW0Z22}. The corpora are indexed with representation and will be retrieved in response to each query based on the relevance scores. The scores are usually calculated based on embedding similarity, such as dot product or cosine similarity. Although dense retrieval systems have developed rapidly, the model performance largely depends supervised text representation learning and relevancy capturing between the query and corpus \cite{DBLP:journals/corr/abs-2211-14876}. Yet, it remains to be a major challenge to properly retrieve when lacking labeled modeling data. Existing work \cite{DBLP:conf/acl/NiACMHCY22, DBLP:conf/emnlp/Ni0LDAMZLHCY22} leveraged pre-trained large encoders (specifically T5 models, \citet{2020t5}) to alleviate the data thirst. However, their proposals still required annotated datasets either by web mining or manual annotation for fine-tuning in order to improve the generalization ability of dual-encoder retrieval models, for example, dealing with out-of-domain data. An alternative solution is to train a dense retrieval on synthetic query-corpus relevance pairs. \cite{DBLP:conf/eacl/MaKYHM21} trains a question generation system on general domain data and applies it to the targeted domain to construct synthetic question-passage data. To save the effort of training a task-specific generation model on general data, like  Natural Questions \cite{DBLP:journals/tacl/KwiatkowskiPRCP19} or MSMARCO \cite{DBLP:conf/nips/NguyenRSGTMD16}, Promptagator \cite{DBLP:conf/iclr/DaiZMLNLBGHC23} proposes to use pre-trained LLMs, like FLAN \cite{DBLP:conf/iclr/WeiBZGYLDDL22}, as a few-shot query generator to build the data for training the dual-encoder. However, the synthetic queries are not directly leveraged at inference, potentially causing gaps between training and inference of dense retrievers \cite{DBLP:conf/acl-deelio/ChoJYP22}. Earlier work, e.g., doc2query \cite{DBLP:journals/corr/abs-1904-08375}, concatenates the generated queries with the corresponding corpus, aiming to enrich the corpus representation with questions that the corpus can potentially answer. An improved version, docTTTTTquery \cite{nogueira2019doc2query} leverages pre-trained T5 models as the expansion model, leading to more relevant synthetic queries and better retrieval performance.

\begin{figure}[htbp]
\centering
\includegraphics[width=0.95\columnwidth]{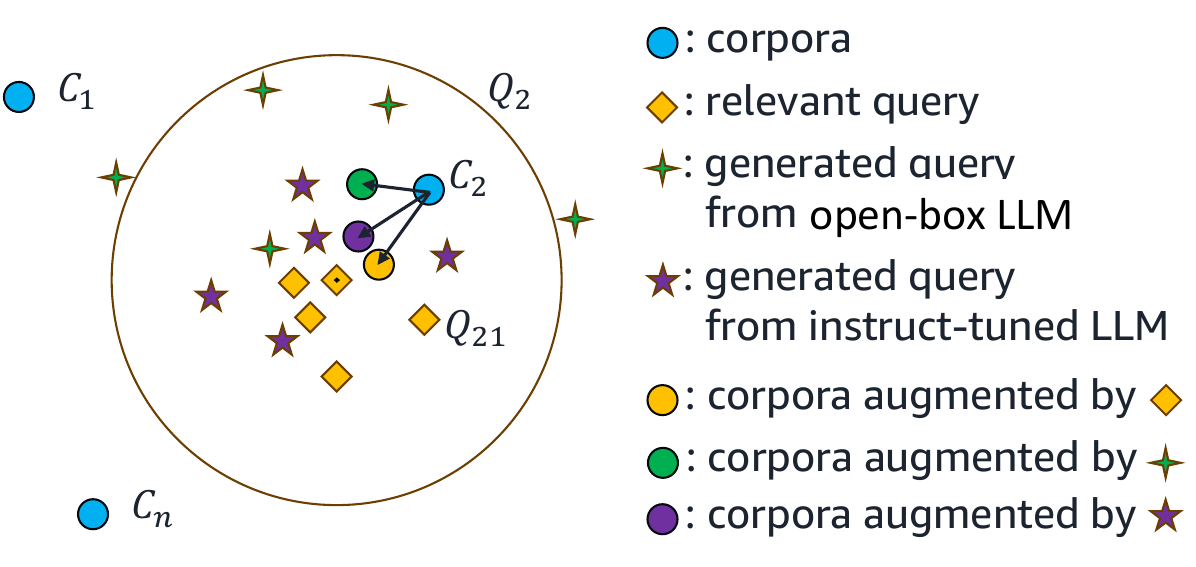} 
\caption{Illustration of the corpus representation augmented by the embedding of relevant queries, synthetic queries generated by open-box LLM and instruct-tuned LLM. }
\label{fig:schematic_diagram}
\end{figure}

\begin{figure*}[htb]
\centering
\includegraphics[width=2\columnwidth]{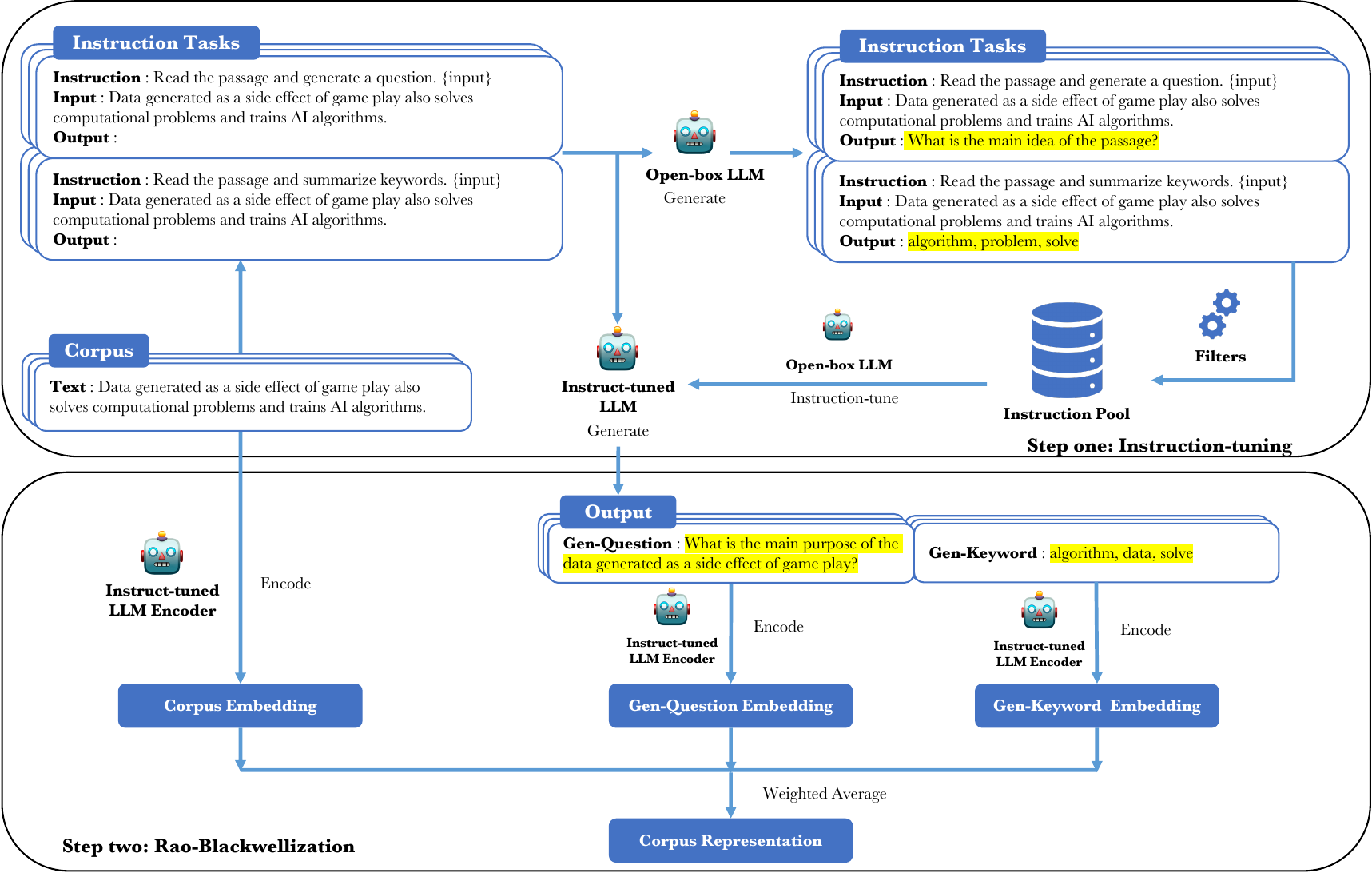} 
\caption{
A high-level overview of Encoder-Decoder corpus representation. In the first instruction-tuning step, given a set of instruction tasks (in our case \textbf{keyword summarization}: ``Read the passage and summarize keywords.'' and \textbf{question generation}: ``Read the passage and generate a question.''), the open-box LLM will generate instruction following examples which are passed through filters for quality control. The filtered examples form an instruction pool and are used to instruction-tune the open-box LLM. In the second Rao-Blackwellization step, by prompting the instruct-tuned LLM using the same instructions as in the first step, synthetic questions and keywords are generated for the corpus. Both the corpus and the generated sequences are encoded by the LLM encoder and the weighted average of their embedding is used as corpus representation.}
\label{fig1}
\end{figure*}

Different from the previous work, we demonstrate directly on the embedding level instead of the text level, that the synthetically generated queries' embeddings can effectively augment the corpus representation (Figure \ref{fig:schematic_diagram}). Here, we propose an unsupervised representation learning approach through self-instructed-tuning leveraging the embedding generation and sequence generation capability of an encoder-decoder LLM. This approach consists of two steps, i.e. self-instructed-learning and Rao-Blackwellization. In the first step, we design two instruction tasks, namely question generation and keyword summarization, to generate synthetic questions and keywords for each given corpus via prompting a pre-trained LLM.  Next, we apply filters to gate the synthetic data quality and instruction-tune the pre-trained LLM (and its variant versions) on the filtered output (Step one in Figure \ref{fig1}). In the second step, we use the instruct-tuned LLM to generate better synthetic questions and keywords following the same instruction prompts as in training. We then obtain the embeddings of the newly generated synthetic questions and keywords and that of the corpus from the instruct-tuned LLM encoder, and take the weighted average as our augmented corpus representation (Step two in Figure \ref{fig1}). 

We consider the corpus representation learning task as a problem of query embedding expectation estimation. Based on the Rao-Blackwell theorem, the crude estimator, corpus embedding, can be improved by taking the conditional expectation given the sufficient statistics, i.e. sample mean of the embedding of their (synthetic) relevant queries and keywords. Thus, we expect combining the raw corpus embedding and synthetic query embedding to achieve better corpus representation. Besides, by aligning instruction-tuning and synthetic query generation, the retrieval model is directly optimized on corpus representation during training. To assess the effectiveness of our proposed method, we compare retrieval method of corpus only embedding with our augmented corpus representation, models with and without instruction-tuning and evaluate against multiple competitive dense retrievers (i.e., mBART \cite{tang2020multilingual}, mDPR \cite{mrtydi, zhang2022best}, T-Systems \cite{tsystem}). Our main contributions are as follows:

\begin{itemize}
  \item We propose a novel unsupervised text representation learning approach for dual-encoder retrieval model by instruction-tuning a pre-trained encoder-decoder using unlabelled corpus. 

  \item We demonstrate our approach of using conditional expectation of the relevant (synthetic) query/keywords embedding the representation of the corpus can be augmented effectively, founded on the Rao-Blackwell theorem. 

  \item We verify the effectiveness of the proposed methods on three English and one German IR datasets measured by NDCG@10, MRR@100, Recall@100. We significantly improve the zero-shot average retrieval performance on all metrics with our unsupervised approach and exceed three competitive supervised dense retrievers, with model of size at least $38\%$ smaller, by $1.96\%$, $4.62\%$, $9.52\%$ absolute on NDCG@10 (Table \ref{tab:sota}).

\end{itemize}

\section{Related Work}

\subsection{Instruction-tuning}  
Tuning pre-trained LLMs with \textit{(natural language instruction, response)} pairs to enhance models’ ability to follow instructions and understand user intention. It is a rising paradigm in NLP to strengthen model’s generalizability on unseen tasks. FLAN \cite{DBLP:conf/iclr/WeiBZGYLDDL22} significantly improves a 137B LLM's zero-shot performance via instruction learning on various NLP datasets with multiple instruction templates. InstructDial \cite{Gupta2022InstructDialIZ} also shows significant zero-shot performance boost in unseen dialogues when applying instruction-tuning to dialogue domain. InstructGPT \cite{Ouyang2022TrainingLM} enhances GPT-3's performance by fine-tuning it on instructions and human feedback collected from OpenAI API. Self-Instruct \cite{DBLP:conf/acl/WangKMLSKH23} fine-tunes the open-box GPT-3 on its own generated instructions and instances which achieved on par performance of InstructGPT.

\subsection{Dense Retrieval Text Representation}
The foundational component of dense retrieval is the text representation. Under dual-encoder framework, it has been a long standing practice such as Sentence-BERT \cite{Reimers2019SentenceBERTSE}, ColBERT \cite{Khattab2020ColBERTEA} to represent query and corpus with encoder only models, e.g., BERT \cite{Devlin2019BERTPO} and RoBERTa \cite{Liu2019RoBERTaAR}. Recently Sentence-T5 \cite{DBLP:conf/acl/NiACMHCY22} demonstrate that encoder-decoder pretrained LLM like T5 can achieve superior performance. Further, representing corpus with single representation may not well model the fine-grained semantic interaction between the queries and corpus. Poly-encoder \cite{Humeau2019PolyencodersAA} and ME-BERT \cite{Luan2020SparseDA} learn multiple representations to better capture the corpus semantics and show significant improvement. Doc2query \cite{DBLP:journals/corr/abs-1904-08375} and docTTTTTquery \cite{nogueira2019doc2query} append generated synthetic queries to the corpus and thus enrich the semantic information.

\section{Method}
We propose an unsupervised text representation learning approach through self-instructed-tuning a pre-trained encoder-decoder LLM. The first step is to generate instruction following responses from an LLM and instruction-tune the LLM itself with filtered quality \textit{(natural language instruction, response)} pairs. The second step is to compute the augmented corpus embedding weighing in synthetic queries' (e.g. questions, keywords) embeddings. Figure \ref{fig1} presents the overall flow of our approach.

\subsection{Problem Scenario}
Denote corpora as $C_1$,$C_2$,...,$C_n$, and their relevant queries as $Q_{11}$, $Q_{12}$,...,$Q_{21}$,..., where queries $Q_{i1}$,$Q_{i2}$,... are relevant to the same corpora $C_i$. For example, $Q_{11}$ can be \texttt{Harry Potter 1} and $Q_{12}$ can be \texttt{Harry Potter and the Philosopher's Stone}, whereas $C_1$ is \texttt{Harry Potter and the Sorcerer's Stone}. $Q_i = {Q_{i1},...,Q_{im}}$

Given a pre-trained encoder-decoder LLM, besides treating the encoder as a text representation model, we consider it as a random variable, where the sample space consists of the range of the possible embedding values, and the corresponding probability measure to each text portion. 
\begin{equation}
\begin{split}
    \texttt{Encoder}(\cdot)&: text \mapsto embedding 
\end{split}
\end{equation}

where the embedding refers to the sentence embedding of the text. 

We assume that an effective encoder maps each group of queries $Q_i$ near a group center in the high-dimensional embedding space and also maps the corresponding $C_i$ to the surrounding area so that $Q_i$ and $C_i$ are well associated. For example, when we have $Q_{21}\in Q_2$ query, the retrieval system will retrieve the $C_2$ corpora which is the closest to the query (Figure \ref{fig:schematic_diagram}).

\noindent
\textbf{Corpus Embedding as an Expectation Estimator}\;\; 
The group center is a comprehensive depiction of the entire group and is indicative to distinguish from other groups. With the pre-trained $\texttt{Encoder}(\cdot)$, the group center is essentially the expected value of the embedding of each group's queries, denoted by $\mathbb{E}(\texttt{Encoder}(Q_i))$. When we use the embedding of the corpus, i.e. $\texttt{Encoder}(C_i)$,  as its representation, we are using it to estimate the group center $\mathbb{E}(\texttt{Encoder}(Q_i))$. This is effective when we don't have any information from the query group. 

\noindent
\textbf{Application of the Rao–Blackwell theorem}\;\; 
Assume we have relevant queries $Q_{i1}$,$Q_{i2}$, ..., $Q_{im}$ for corpus $C_i$. Then $\frac{1}{m}\sum_{j=1}^m\\\texttt{Encoder}(Q_{ij})$ is a sufficient statistics to estimate $\mathbb{E}(\texttt{Encoder}(Q_i))$.

According to Rao–Blackwell Theorem: If $g(\mathbf{X})$ is any kind of estimator of a parameter $\theta$, then the conditional expectation of $g(\mathbf{X})$ given $T(\mathbf{X})$, namely $\mathbb{E}(g(x)|T(x))$, where $T$ is a sufficient statistic, is typically a better estimator of $\theta$, and is never worse. Plug in Equation (\ref{eq:rao_black}), we get an improved estimator for $\mathbb{E}(\texttt{Encoder}(Q_i))$, which is $\mathbb{E}(\texttt{Encoder}(C_i)|\frac{1}{m}\sum_{j=1}^m \texttt{Encoder}(Q_{ij}))$.

\begin{equation}
    \begin{split}
        g(x)&=\texttt{Encoder}(C_i) \\
        T(x)&=\frac{1}{m}\sum_{j=1}^m \texttt{Encoder}(Q_{ij}) \\
        \theta&=\mathbb{E}(\texttt{Encoder}(Q_i)) 
    \end{split}
\label{eq:rao_black}
\end{equation}

With some regularity assumptions, e.g., $C_i\in Q_i$ and $C_i=Q_{i1}$, the conditional expectation can be written as 
\begin{equation}
\begin{split}
    &\mathbb{E}(\texttt{Encoder}(C_i)|\frac{1}{m}\sum_{j=1}^m \texttt{Encoder}(Q_{ij})) \\
    &=\frac{1}{m}\sum_{j=1}^m \texttt{Encoder}(Q_{ij}) \\
    &=\frac{1}{m}\texttt{Encoder}(C_i) + \frac{1}{m}\sum_{j=2}^m \texttt{Encoder}(Q_{ij})
\end{split}
\label{eq:conditional_expectation}
\end{equation}
We expect to achieve better performance with this formula for corpus representation. An intuitive understanding is that it gets closer to the relevant queries' embedding in the vector space (Figure \ref{fig:schematic_diagram}).

\subsection{Synthetic Query Generation}
Obtaining a comprehensive set of labeled queries is labor-intensive and costly, especially in low resource setting. LLMs have built their reputation as generative AI models and are capable of following well designed instructions. Not only can the model generate text, but it also can output the generation probability of the text. We denote the generation model by $\texttt{LLM}(\cdot)$, then the generation can be written as,
\begin{equation}
    \hat{Q}_{ij}, \hat{P}(\hat{Q}_{ij}) = \texttt{LLM}(\texttt{Instruction}+C_i)
\label{eq:instruction}
\end{equation}
where $\hat{Q}_{ij}$ is the generated query and $\hat{P}(\hat{Q}_{ij})$ is the generation probability. The instruction is a pre-defined generation task, for example ``write a question for'' or ``what are the keywords of''. 

\subsection{Corpus Representation}
Plug in the generated synthetic queries, let $R(C_i)$ denote the final representation of corpora $C_i$, the Equation (\ref{eq:conditional_expectation}) becomes a weighted average of original corpora embedding and its synthetic query embedding,
\begin{equation}
\begin{split}
    R(C_i)\;\hat{=}\;&w_0\texttt{Encoder}(C_i)\;+\\ 
    &(1-w_0)\sum_j \hat{P}(\hat{Q}_{ij})\texttt{Encoder}(\hat{Q}_{ij})
\end{split}
\label{eq:corpora_representation}
\end{equation}
where $w_0$ is a hyper-parameter that is tuned on a subset of test queries. Equation (\ref{eq:corpora_representation}) is our proposed corpus representation for the dual-encoder retrieval system. Note that we can generate different types of synthetic queries in Equation (\ref{eq:instruction}) using various instructions, and we can generate multiple sequences for each instruction by adopting decoding strategies such as beam search. We can also improve the quality of the generated queries through instruction-tuning as follows.

\subsection{Instruction-Tuning the LLM}
While LLM demonstrates reasonable text generation capabilities, its ability to follow specific instructions can be honed. Instruction-tuning focuses on training a model to precisely follow the provided instructions. 

As we don't have the query-corpora labeled data, we propose to self-instructed-tuning the LLM on its self-generated quality (i.e. gated) responses following given instructions to enhance synthetic queries generation relevance. This approach has demonstrated its effectiveness \cite{DBLP:conf/acl/WangKMLSKH23}. The instruct-tuned LLM is then used to prepare the synthetic queries for the corpus representation augmentation as in Equation (\ref{eq:instruct_tuned}).



\begin{equation}
\begin{split}
    &\hat{Q}_{ij}, \hat{P}(\hat{Q}_{ij}) = \\
    &\texttt{InstructTunedLLM}(\texttt{Instruction}+C_i)
\end{split}
\label{eq:instruct_tuned}
\end{equation}
We use the same instructions across the entire framework, including generation and training. Figure \ref{fig:schematic_diagram} shows a schematic diagram that although the generated queries from an open-box pre-trained LLM may not effectively enrich the corpora, after instruction-tuning, the generated synthetic queries become more relevant and the corpus representation can be improved consequently. 

\section{Experiments}

\subsection{Datasets} 
In this work, we tested four IR datasets where the summary of the database is shown in Table \ref{tab:dataset}. \textit{\textbf{English:}} (1) NFCorpus \cite{boteva2016} has automatically extracted relevance judgments for medical documents. (2) SciFact \cite{wadden-etal-2020-fact} consists of expert-annotated scientific claims with abstracts and rationales. (3) SCIDOCS \cite{cohan-etal-2020-specter} has seven document-level tasks from citation prediction, document classification, and recommendation. \textit{\textbf{German:}} (4) GermanQuAD \cite{DBLP:journals/corr/abs-2104-12741} has the relevant information for high complex German QA with a large size of corpora. Due to computation resource limits, we downsample the corpus in SCIDOCS and GermanQuAD datasets, where we ensure the downsampled corpus include all relevant corpus for test queries. Note that such downsampling does not prevent us from fairly comparing the zero-shot retrieval efficacy of our approach with open-box LLMs because all experiments are performed under the same data setting. To help the encoder capture the fine-grained semantic interaction between queries and corpus, we divide each corpora into multiple sentences using the PunktSentenceTokenizer \footnote{\url{https://www.nltk.org/api/nltk.tokenize.PunktSentenceTokenizer.html}} from nltk package and use the sentence level multi representation for the corpora, meaning that if any of the sentence is retrieved, the passage is retrieved.

\begin{table}
\centering
\caption{Details of datasets used. The size of corpus is downsampled to 15K in SCIDOCS and 10K in GermanQuAD. Filtered Queries: Generated synthetic queries from FLAN-T5-Large with filtering.}
\scalebox{0.75}{
\begin{tabular}{c|c|c|c|c} 
\toprule
\textbf{\textbf{Dataset}} & \textbf{\textbf{Language}} & \begin{tabular}[c]{@{}c@{}}\textbf{Test}\\\textbf{Queries}\end{tabular}  & \textbf{Corpus} & \begin{tabular}[c]{@{}c@{}}\textbf{Filtered}\\\textbf{Queries}\end{tabular}  \\ 
\midrule
NFCorpus                  & English                    & 323              & 3.6K            & 5.9K                                                                           \\ 
SciFact                   & English                    & 300              & 5.1K            & 8.2K                                                                           \\ 
SCIDOCS                   & English                    & 1K               & 25.6K           & 29.4K                                                                          \\ 
GermanQuAD                & German                     & 2K               & 2.8M            & 17.5K                                                                          \\
\bottomrule
\end{tabular}}
\label{tab:dataset}
\end{table}

\begin{table}
\centering
\caption{Open-box encoder-only average performance for passage level and sentence level indexing. Model is FLAN-T5. $\spadesuit$: NDCG@10. $\clubsuit$: MRR@100 $\heartsuit$: Recall@100.}
\scalebox{0.75}{
\begin{tabular}{l|c|ccc} 
\toprule
\multicolumn{1}{c|}{\textbf{\textbf{\textbf{\textbf{Index}}}}} &\diagbox[font=\small]{\textbf{Model}}{\textbf{Metric}} & $\spadesuit$   & $\clubsuit$    & $\heartsuit$    \\ 
\midrule
\multirow{2}{*}{Passage}                                       & Base                                                                                                                       & 5.79           & 8.02           & 22.75           \\
                                                               & Large                                                                                                                      & 8.78           & 10.63          & 32.43           \\ 
\midrule
\multirow{2}{*}{Sentence}                                      & Base                                                                                                                       & 22.02          & 25.96          & 43.54           \\
                                                               & Large                                                                                                                      & \textbf{23.15} & \textbf{26.53} & \textbf{46.18}  \\
\bottomrule
\end{tabular}
}
\label{tab:sentence}
\end{table}

\begin{table*}
\centering
\caption{Comparison of performances according to instruction-tuning.}
\scalebox{0.7}{
\begin{tabular}{c|c|ccc|ccc|ccc|ccc|ccc} 
\toprule
\multirow{3}{*}{\begin{tabular}[c]{@{}c@{}}\textbf{Instruction-}\\\textbf{tuning}\end{tabular}} & \textbf{Dataset}      & \multicolumn{3}{c|}{NFCorpus} & \multicolumn{3}{c|}{SciFact} & \multicolumn{3}{c|}{SCIDOCS} & \multicolumn{3}{c|}{GermanQuAD  } & \multicolumn{3}{c}{\textbf{Avg }}  \\ 
\cline{2-17}
                                                                                                          & \diagbox[font=\small]{\textbf{Model}}{\textbf{Metric}} & $\spadesuit$ & $\clubsuit$ & $\heartsuit$             & $\spadesuit$ & $\clubsuit$ & $\heartsuit$            & $\spadesuit$ & $\clubsuit$ & $\heartsuit$               & $\spadesuit$ & $\clubsuit$ & $\heartsuit$                 & $\spadesuit$ & $\clubsuit$ & $\heartsuit$                       \\ 
\midrule
\multirow{2}{*}{No}                                                                           & Base                  & 12.15 & 26.58 & 15.8          & 29.62 & 28.54 & 66.28        & 6.4  & 13.39 & 17.74           & 49.41 & 45.82 & 83.17             & 24.39 & 28.58 & 45.75              \\ 
                                                                                                          & Large                 & 10.42 & 23.41 & 14.6          & 30.66 & 28.84 & 71.45        & \textbf{7.22} & 14.07 & 22.1            & 50.82 & 47.24 & 83.61             & 24.78 & 28.39 & 47.94              \\ 
\midrule
\multirow{2}{*}{Yes}                          & Base                  & \textbf{12.3}  & 27.02 & \textbf{16.24}         & 30.72 & 29.59 & 65.06        & 6.04 & 12.7  & 16.5            & 52.39 & 48.54 & 84.44             & 25.36 & 29.46 & 45.56              \\ 
                                                                                                          & Large                 & 11.91 & \textbf{27.04} & 15.85         & \textbf{32.03} & \textbf{29.92} & \textbf{73.21}        & 7.16 & \textbf{14.57} & \textbf{22.39}           & \textbf{55.49} & \textbf{51.99} & \textbf{86.79}             & \textbf{26.65} & \textbf{30.88} & \textbf{49.56}              \\
\bottomrule
\end{tabular}}
\label{tab:instruction_effect}
\end{table*}

\begin{table*}
\centering
\caption{Comparison with SOTA. Tuned: Model with instruction-tuning.}
\scalebox{0.7}{
\begin{tabular}{c|c|ccc|ccc|ccc|ccc|ccc} 
\toprule
\multicolumn{2}{c|}{\textbf{Dataset}} & \multicolumn{3}{c|}{NFCorpus}                    & \multicolumn{3}{c|}{SciFact}                     & \multicolumn{3}{c|}{SCIDOCS}                    & \multicolumn{3}{c|}{GermanQuAD}                  & \multicolumn{3}{c}{\textbf{Avg }}                 \\ 
\cline{1-2}\cmidrule{3-17}
Model               & \diagbox[font=\small]{\textbf{Size}}{\textbf{Metric}}            & $\spadesuit$   & $\clubsuit$    & $\heartsuit$   & $\spadesuit$   & $\clubsuit$    & $\heartsuit$   & $\spadesuit$  & $\clubsuit$    & $\heartsuit$   & $\spadesuit$   & $\clubsuit$    & $\heartsuit$   & $\spadesuit$   & $\clubsuit$    & $\heartsuit$    \\ 
\cline{1-2}\cmidrule{3-17}
mDPR                & 177M             & 8.30           & 19.19          & 11.57          & 23.46          & 21.87          & 58.94          & 4.75          & 10.26          & 15.98          & \textbf{57.09} & \textbf{53.19} & \textbf{87.67} & 23.40          & 26.13          & 43.54           \\ 
\cline{1-2}\cmidrule{3-17}
T-Systems           & 278M             & \textbf{15.32} & \textbf{29.14} & \textbf{17.05} & 25.32          & 23.74          & 59.29          & \textbf{8.38} & \textbf{17.64} & \textbf{23.82} & 33.93          & 30.95          & 64.14          & 20.74          & 25.37          & 41.08           \\ 
\cline{1-2}\cmidrule{3-17}
mBART-Large         & 331M             & 1.87           & 5.87           & 4.56           & 23.85          & 22.52          & 52.53          & 3.58          & 7.82           & 12.69          & 34.06          & 31.47          & 63.31          & 15.84          & 16.92          & 33.27           \\ 
\cline{1-2}\cmidrule{3-17}
Tuned-FLAN-T5-Base  & 109M             & 12.30          & 27.02          & 16.24          & \textbf{30.72}          & \textbf{29.59}          & \textbf{65.06}          & 6.04          & 12.70          & 16.50          & 52.39          & 48.54          & 84.44          & \textbf{25.36}          & \textbf{29.46}          & \textbf{45.56}           \\ 
\hhline{=================}
Tuned-FLAN-T5-Large & 341M             & 11.91          & 27.04          & 15.85          & \textit{\textbf{32.03}} & \textit{\textbf{29.92}} & \textit{\textbf{73.21}} & 7.16          & 14.57          & 22.39          & 55.49          & 51.99          & 86.79          & \textit{\textbf{26.65}} & \textit{\textbf{30.88}} & \textit{\textbf{49.56}}  \\
\bottomrule
\end{tabular}
}
\label{tab:sota}
\end{table*}

\subsection{Baseline}
We compare between the corpus-only representation and our proposed augmented corpus representation in zero-shot experiments under the dual-encoder framework. To obtain the representation of a sequence from the encoder, we perform mean aggregation over the last hidden state of each token following \cite{DBLP:conf/acl/NiACMHCY22}. We measure the relevance between query and corpus using cosine similarity. 

To understand the superiority of our approach, we compare with three different SOTA models: (1) mDPR \cite{mrtydi, zhang2022best} is a variation of DPR model \cite{karpukhin-etal-2020-dense} which replaces BERT to multilingual BERT \cite{Devlin2019BERTPO} to support non-English languages for retrieval tasks. (2) T-Systems \cite{tsystem} is developed for computing sentenc embeddings for English and German texts. It uses a XLM-RoBERTa \cite{DBLP:journals/corr/abs-1911-02116} and is fine-tuned with English-German datasets. (3) mBART-Large \cite{tang2020multilingual} is a multilingual Sequence-to-Sequence generation model. It supports 50 languages and we consider it for comparison in same model structure (i.e. encoder-decoder). Lastly, we compare with docTTTTTquery \cite{nogueira2019doc2query} to understand the effectiveness of our corpus representation augmentation.

\subsection{Encoder-Decoder Models}
T5 is an encoder-decoder model pre-trained on a combination of unsupervised and supervised tasks, where each task is transformed into a text-to-text format \cite{2020t5}. FLAN-T5 is an enhanced version of T5 fine-tuned on a mixture of tasks \cite{DBLP:conf/iclr/WeiBZGYLDDL22}. Considering that these types of models are open source, offer various sizes, support English and German, and have an encoder-decoder architecture, we leverage the FLAN-T5-Base and Large models in our experiments. 

\subsection{Instruction Query Generation}
For instruction query generation and instruction-tuning, we consider two types of instructions (i.e. keyword summarization and question generation) as shown in Figure \ref{fig1}. We also develop a filter to improve the quality of generated instructions. If the task is keyword summarization, the number of keywords should be smaller than the half number of sentences in corpus. If it's question generation, the generated sequence should end with a question mark. The filter is simple, leaving room for further improvement. The numbers of the filtered synthetic queries are shown in Table \ref{tab:dataset}.

    



\subsection{Hyperparameter Setting}
When performing instruction-tuning, we use the same hyperparameter setting for all the models. Specifically, we use the AdaFactor optimizer with learning rate 0.0001, batch size 16, and the number of epochs 30. Early stopping is performed when the validation loss shows no improvement for five consecutive epochs. 


When generating queries using FLAN-T5 models, we only consider one returned sequence for each instruction and assume they are equally important. We denote the generated question and keywords as $\hat{question_i}$ and $\hat{keywords_i}$. We tested the multiple weighting methods for corpus representation where the best approach is giving the weight on the original corpus as $w_0=0.6$, so that each of $\hat{question_i}$ and $\hat{keywords_i}$ has the weight $0.2$. Thus, the corpus representation is:
\begin{equation}
\begin{split}
    &\texttt{R}(C_i)=0.6\times\texttt{Encoder}(C_i)+0.2\;\times \\
    &(\texttt{Encoder}(\hat{question_i})+\texttt{Encoder}(\hat{keywords_i}))
\end{split}
\label{eq:weight}
\end{equation}

\section{Results and Discussion}
\subsection{Corpora vs Sentence Indexing}

We evaluate whether the sentence level multi-representation can capture the semantic interaction between the corpora and the query. Results for FLAN-T5 models using encoder-only representation are shown in Table \ref{tab:sentence}. The sentence level multi-representation embedding technique outperforms the corpora level single representation by a large margin across all datasets. As the model size increases, the performance also gets better. Note that our approach uses no labeled data to achieve on par performance as SOTA models, and sentence level indexing is a way we do for chunking. According to the promising empirical results, we will apply the sentence level multi-representation technique to all the following experiments.



\begin{figure*}[htb]
\centering
\includegraphics[width=0.99\linewidth]{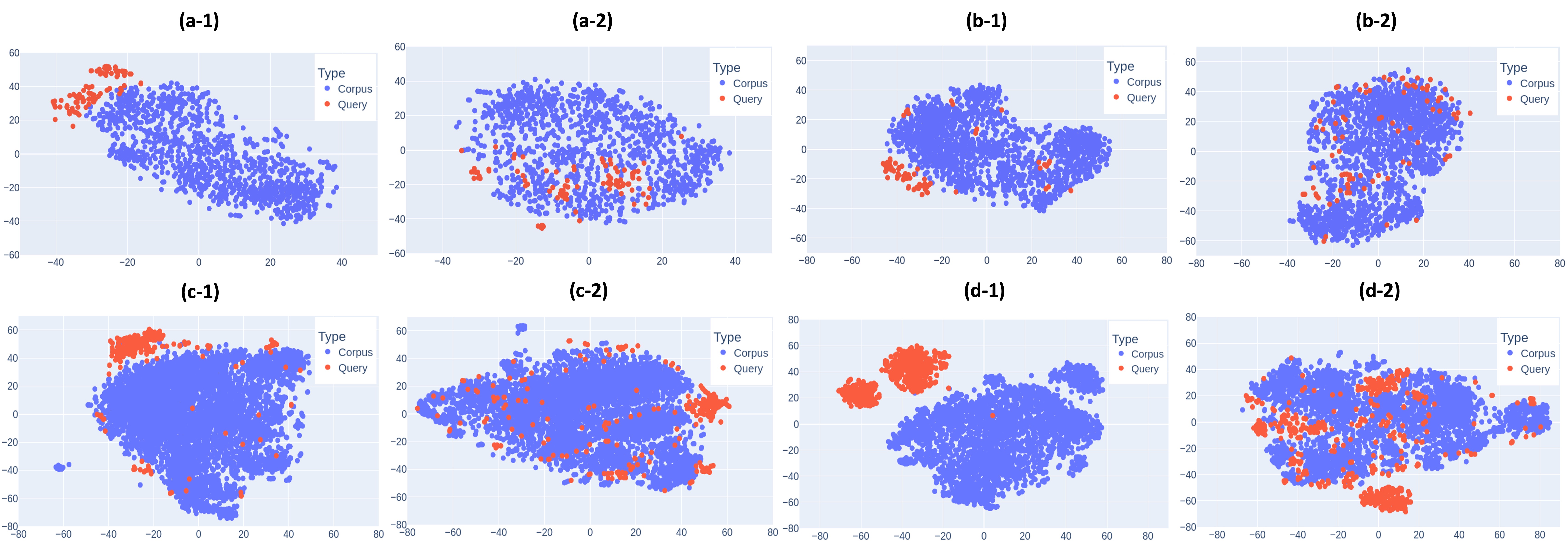} 
\caption{t-SNE distributions for corpus representation generated from FLAN-T5-Large. (a-d) NFCorpus, SciFact, SCIDOCS, GermanQuAD. (1-2) Original corpus, Weighted corpus with synthetic queries after instruction-tuning.}
\label{fig:tsne}
\end{figure*}

\subsection{Overall Results}
Table \ref{tab:instruction_effect} describes the performance of FLAN-T5 models regarding instruction-tuning. Overall, we can mostly find the improvements of performances in all metrics after instruction-tuning, except for SCIDOCS. This is mainly because the quality of generated queries after instruction-tuning are proper and detailed (Table \ref{tab:example}), and also each synthetic query is less overlapped which makes the corpora distinguishable. The influence of instruction-tuning is greater in larger model since it can have better generation capability and be more affected by fine-tuning with instructions. 

Table \ref{tab:sota} compares our approach and SOTA models in zero-shot scenarios. To clarify, FLAN-T5-Base has similar size as other SOTA models which can be considered as a fair comparison. First of all, instruct-tuned FLAN-T5-Base shows the boosted averaged results than other SOTA models which reveals the prowess of our approach. Considering a larger model (i.e. Tuned-FLAN-T5-Large) enhances the performance further. Thus, our suggested approach is consistently applicable in different size of models where the larger one promises the better performances.

\subsection{Ablation Study}

\noindent
\textbf{Optimal Corpus Representation}\;\; 
From our findings, new corpus representation based on synthetic queries from instructions is useful to improve the retrieval performances. To define the optimal weights in corpus representation, we investigate the four different weight methods: (1) Equal: Giving the equal weights for corpus and generated synthetic queries (i.e. keyword, question). (2) Manual: It is same as Equation (\ref{eq:weight}). (3) BERTScore: Giving the weights based on BERTScore (F1) with BERT-Base-Multilingual-Cased model \cite{DBLP:journals/corr/abs-1810-04805}. Equation (\ref{eq:bert}) shows the details of it. (4) BERTScore$_{Softmax}$: Similar as BERTScore but including the Softmax.

\begin{equation}
\begin{split}
    &X: \hat{keywords_i}, \hat{question_i} \\
    &\texttt{BERT}_X:\;\texttt{BERTScore}\;\text{between}\;X\;\text{and}\;C_i \\
    &\texttt{Denominator}=1+\texttt{Sum}(\texttt{BERT}_X) \\
    &\texttt{Weight}_X = \frac{\texttt{BERT}_X}{\texttt{Denominator}}, \\
    &\texttt{Weight}_{C_i} = \frac{1}{\texttt{Denominator}}
\end{split}
\label{eq:bert}
\end{equation}

Table \ref{tab:representation} shows the overall performances according to the different weight approaches in corpus representation. First of all, the equal weight approach shows the worst performance which confirms that the corpus basically contains the most relevant information for queries which should be weighted more. Also, extracted keywords and questions mostly have the essential contexts but partial information of corpus which is not enough to include the semantic meaning of corpus. Thus, manual weighting with emphasis on corpus promises the better result than BERTScore approaches. 

\begin{table*}
\centering
\caption{Example of synthetic queries according to the instruction-tuning. FLAN-T5-Large is used for generating the examples.}
\scalebox{0.7}{
\begin{tabular}{l|l|l|l} 
\toprule
\multicolumn{1}{c|}{\textbf{Corpus}}                                                                                                                                                                                                                                                                                                                                                                                                                   & \multicolumn{1}{c|}{\textbf{Instruction Type}} & \multicolumn{1}{c|}{\textbf{Open-box}}                                                                                                   & \multicolumn{1}{c}{\textbf{Instruct-tuned}}                                                               \\ 
\midrule
\begin{tabular}[c]{@{}l@{}}Semantic Space is a pervasive computing infrastructure that exploits\\semantic Web~technologies to support explicit representation, \\expressive querying, and flexible reasoning~of contexts in smart spaces.\end{tabular}                                                                                                                                                                                                 & Keyword                            & \begin{tabular}[c]{@{}l@{}}context, support, \\query, semantic, \\space\end{tabular}                                                          & semantic space                                                                                               \\ 
\midrule
\begin{tabular}[c]{@{}l@{}}Fluorometric titration of E. coli single-stranded DNA binding protein with various\\RNAs showed that the protein specifically and cooperatively binds to its own mRNA.\\The binding inhibited in vitro expression of ssb and bla but not nusA. This inhibition\\takes place at a physiological concentration of SSB. The function of the protein in\\gene regulation is discussed.\end{tabular}                             & Keyword                            & \begin{tabular}[c]{@{}l@{}}The single-stranded\\DNA binding protein(SSB)\\specifically and cooperatively\\binds to its own mRNA.\end{tabular} & mRNA, protein, titration                                                                                     \\ 
\midrule
\begin{tabular}[c]{@{}l@{}}This paper describes an aggregation and correlation algorithm used in~the\\design and implementation~of an intrusion-detection console built on~top of\\the Tivoli Enterprise Console (TEC). The aggregation~and correlation algorithm\\aims at acquiring intrusion-detection alerts and relating them together to expose\\a more condensed view of the security issues raised by intrusion-detection systems.\end{tabular} & Question                           & \begin{tabular}[c]{@{}l@{}}What is the purpose\\of the paper?\end{tabular}                                                                    & \begin{tabular}[c]{@{}l@{}}What is the purpose of the\\aggregation and correlation\\algorithm?\end{tabular}  \\ 
\midrule
\begin{tabular}[c]{@{}l@{}}ESC is to create an inventory of cardiovascular disease registries and\\a task force on data standardization\end{tabular}                                                                                                                                                                                                                                                                                                   & Question                           & \begin{tabular}[c]{@{}l@{}}What is the purpose\\of the task force?\end{tabular}                                                               & \begin{tabular}[c]{@{}l@{}}What is the purpose of the\\ESC?\end{tabular}                                     \\
\bottomrule
\end{tabular}
}
\label{tab:example}
\end{table*}

\begin{table}
\centering
\caption{Different weight methods for corpus representation. Model is based on FLAN-T5.}
\scalebox{0.75}{
\begin{tabular}{c|c|ccc} 
\toprule
\begin{tabular}[c]{@{}c@{}}\textbf{\textbf{\textbf{\textbf{Corpus}}}}\\\textbf{\textbf{\textbf{\textbf{Weights}}}}\end{tabular} & \diagbox[font=\small]{\textbf{Model}}{\textbf{Metric}} & $\spadesuit$ & $\clubsuit$ & $\heartsuit$   \\ 
\hline
\multirow{2}{*}{N/A}                                                                                                            & Base                        & 22.02 & 25.96 & 43.54    \\
                                                                                                                                & Large                       & 23.15 & 26.53 & 46.18  \\ 
\hline
\multirow{2}{*}{Equal}                                                                                                          & Base                        & 18.25 & 21.98 & 38.76    \\
                                                                                                                                & Large                       & 17.92 & 21.60 & 39.93    \\ 
\hline
\multirow{2}{*}{Manual}                                                                                                         & Base                        & 24.39 & \textbf{28.58} & 45.75    \\
                                                                                                                                & Large                       & \textbf{24.78} & 28.39 & \textbf{47.94}    \\ 
\hline
\multirow{2}{*}{BERTScore}                                                                                                      & Base                        & 22.35 & 26.12 & 43.58    \\
                                                                                                                                & Large                       & 21.97 & 25.53 & 45.2     \\ 
\hline
\multirow{2}{*}{BERTScore$_{Softmax}$}                                                                                         & Base                        & 20.13 & 23.64 & 40.68    \\
                                                                                                                                & Large                       & 19.52 & 23.12 & 42.72    \\
\bottomrule
\end{tabular}
}
\label{tab:representation}
\end{table}

\noindent
\textbf{Effectiveness of Instruction-tuning}\;\; 
Table \ref{tab:example} gives the examples of generated synthetic queries. In keyword summarization, open-box extracts the ambiguous and meaningless words (first example) or a simple copy of sentence (second example) as keywords while instruction-tuning helps to observe the whole corpus to extract the core keywords. For question generation, open-box generates the general (third example) or unanswerable questions (fourth example) while instruction-tuning gives the detailed and suitable questions which can be accountable by the specific corpus. 

Figure \ref{fig:tsne} shows the distributions of embeddings of corpora and test queries based on FLAN-T5-Large. Overall, the weighted corpus representation and instruction-tuning spread out the corpora embeddings to make them distinguishable. Also, it helps to locate the test queries closer to the corpora. Thus, our approach helps to integrate the crucial and detailed synthetic queries for corpus representation which helps to generate the unique corpora to achieve the enhanced retrieval performances.

\noindent
\textbf{Effectiveness of Corpus Representation Augmentation}\;\; 
We compare with other corpus representation augmentation, docTTTTTquery \cite{nogueira2019doc2query}, to validate our corpus augmentation. Here, we follow the default strategy of docTTTTTquery: top-10 with 40 predictions appending on corpus. According to Table \ref{tab:corpus_aug}, we demonstrate significant improvement via our approach - embedding level augmentation with representations from self-instructed-tuned model. Based on this finding, we can confirm that augmenting representation on embedding level is more effective than on input text level with concatenation as docTTTTTquery, and our self-instructed-tuned model performs better than their supervised representation generation model.

\begin{table}
\centering
\caption{Different corpus representation augmentation. Model is based on FLAN-T5. Note that we evaluated on English datasets.}
\scalebox{0.85}{
\begin{tabular}{c|c|ccc} 
\toprule
\textbf{Model}                 &  \diagbox[font=\small]{\textbf{Method}}{\textbf{Metric}} & $\spadesuit$ & $\clubsuit$ & $\heartsuit$  \\ 
\hline
\multirow{2}{*}{Base}  & docTTTTTquery                                                                                                              & 4.94         & 10.12       & 14.33         \\
                               & Our approach                                                                                                        & \textbf{16.35}        & \textbf{23.1}       & \textbf{32.6}         \\ 
\hline
\multirow{2}{*}{Large} & docTTTTTquery                                                                                                              & 6.87         & 11.81       & 19.93         \\
                               & Our approach                                                                                                         & \textbf{17.03}        & \textbf{23.84}        & \textbf{37.15}         \\
\bottomrule
\end{tabular}}
\label{tab:corpus_aug}
\end{table}

\section{Conclusion}
In our research, we propose the unsupervised text representation learning technique through self-instructed-tuning encoder-decoder LLMs. Based on the Rao-Blackwell theorem, we leverage the embeddings of synthetically generated queries (i.e. questions and keywords) to augment the corpus representation for the dual-encoder retrieval framework. In zero-shot experiments, our proposed corpus representation consistently improves the performance over encoder-only corpus representation. Even if the open-box LLM was not pre-trained on retrieval task and there is no labeled modeling data, after fine-tuning with our approach it exceeds the SOTA models across different datasets, presenting the high effectiveness and data efficiency of our method in retrieval tasks.

In future work, we plan to explore our proposed method on separate encoder and decoder models.




\bibliographystyle{ACM-Reference-Format}
\bibliography{Main}

\end{document}